# Deep Learning Models for Water Stage Predictions in South Florida

[1]Jimeng Shi[1†], Zeda Yin[2†], Rukmangadh Sai Myana[1], Khandker Ishtiaq[3], Anupama John[3], Jayantha Obeysekera[3], Arturo S. Leon[2‡], Giri Narasimhan[1‡]

[1]Knight Foundation School of Computing and Information Sciences, Florida International University
[2]Department of Civil and Environmental Engineering, Florida International University
[3]Sea Level Solutions Center, Florida International University

{jshi008, zyin005, rmyan001, arleon, kishtiaq, anujohn, jobeysek, giri}@fiu.edu

[†]Equal Contributions, [‡]Corresponding Author

**ABSTRACT**

Simulating and predicting the water level/stage in river systems is essential for flood warnings, hydraulic operations, and flood mitigations. Physics-based detailed hydrological and hydraulic computational tools, such as HEC-RAS, MIKE, and SWMM, can be used to simulate a complete watershed and compute the water stage at any point in the river system. However, these physics-based models are computationally intensive, especially for large watersheds and for longer simulations, since they use detailed grid representations of terrain elevation maps of the entire watershed and solve complex partial differential equations (PDEs) for each grid cell. To overcome this problem, we train several deep learning (DL) models for use as surrogate models to rapidly predict the water stage. A portion of the Miami River in South Florida was chosen as a case study for this paper. Extensive experiments show that the performance of various DL models (MLP, RNN, CNN, LSTM, and RCNN) is significantly better than that of the physics-based model, HEC-RAS, even during extreme precipitation conditions (i.e., tropical storms), and with speedups exceeding 500x. To predict the water stages more accurately, our DL models use both measured variables of the river system from the recent past and covariates for which predictions are typically available for the near future.

## 1. INTRODUCTION

Floods are seasonal in many areas but are mostly rare natural events caused by increased flow rate and water stage in a river, resulting in temporarily submerging of low-lying areas along its banks (Parhi, 2018). Floods result in commercial losses and dangers to human life (Grothmann & Reusswig, 2006; Hirabayashi & Kanae, 2009; Alexander et al., 2019) and could contribute to environmental and public health risks (Okaka & Odhiambo, 2018; Rivett et al., 2022). Thus, knowing the time and location of floods ahead of time is critical for hydraulic structure operators to make informed decisions for flood mitigation promptly (Kusler, 2004; Leon et al, 2020) and for citizens and local governments to be better prepared for potential flood outcomes. Higher extremes in rainfall due to climate change will likely increase the frequency and volume of floods (Seneviratne et al., 2022). Currently, physics-based models are available to simulate river floods, including Hydrologic Engineering Center's River Analysis System (HEC-RAS), MIKE developed by the Danish Hydraulic Institute (DHI) company, and Storm Water Management Model (SWMM). However, thousands of simulations of these physics-based models may be required to find the optimal flow release on the control of hydraulic structures (e.g., gates, pumps, dams, reservoirs) to mitigate the floods (Leon et al., 2020; Leon et al., 2021). In real-time applications, the use of conventional physics-based models is infeasible since they are extremely time-consuming by simulating each cross-section in the river systems, especially for large complex domains with multiple driving factors, e.g., heavy rain, strong winds, and storm surges. Therefore, there is a critical need for a surrogate model that can rapidly and accurately predict the water stages at specific locations along the river.

Several studies have explored the application of various machine learning (ML) and deep learning (DL) models for a range of purposes. The artificial neural network (ANN) has been applied to forecast river flow in the Apalachicola River (Huang et al., 2004) and net inflow volumes into Lake Okeechobee (Obeysekera et al., 2006). Huang et al., 2004 compared the results of the forecasted flow rates from the ANN model with those from a traditional autoregressive integrated moving average (ARIMA) forecasting model. Sadler et al. (2018) used a Random Forest model to predict flood severity in Norfolk, Virginia, USA. Given extensive environmental data (i.e., rainfall, tide, groundwater table level, and wind conditions) as input, they focused on a classification task by predicting the number of flood reports. Furthermore, Sapitang et al. (2020) utilized four ML models to enhance daily forecasts of reservoir water levels and reported the ML intra-model comparison. The ML models used in their work include Boosted Decision Tree Regression, Decision Forest Regression, Bayesian Linear Regression, and Neural Network Regression. Similarly, Ayus et al. (2023) employed Random Forest, XGBoost, and bidirectional-LSTM (long short-term memory) network models on 30 years of data to predict daily water levels in Jezioro Kosno Lake in Poland.

As shown above, ML and DL approaches have been widely applied to model environmental data. However, we noted the following limitations. First, most efforts directly tried off-the-shelf ML models, which may be limiting their ability to model increasingly complex data sets. Second, most only reported errors in comparison with observed data or simply provided ML intra-model comparison (Sapitang et al., 2020), but did not report the comparison with the performance of current physics-based models. Third, some of them (Sapitang et al., 2020; Ayus et al., 2023) worked on coarse time granularity, e.g., daily, weekly, or even monthly. More importantly, none of these methods incorporated future covariates into their models, especially the covariates that can be estimated or determined in advance, e.g., rainfall information and control settings of hydraulic structures.

In this work, we aim to address the limitations above. Our main contributions are as follows. We show that DL models perform better than physics-based models such as HEC-RAS for prediction tasks. While the methods are comparable in prediction accuracy, DL methods are orders of magnitude faster than their physics-based counterparts, opening the possibility for real-time flood management. They also outperform simpler regression and standard ML methods. To fine-tune our understanding of the applicability of DL methods to this type of data, we applied multiple DL models (MLP, RNN, CNN, and LSTM) to predict the water stages. We also proposed a combinational model, RCNN, by combining recurrent neural networks (RNN) and convolutional neural networks (CNN) in series. We concluded that the differences in performance between the DL models were not statistically significant. We conducted extensive experiments to study the impact of varying forecast lead times. We proved that RCNN marginally outperforms other DL models over forecast lead times. We demonstrated that incorporating future covariates into DL models can lead to improved predictions, even if they are low in accuracy. To simulate low accuracy forecasts, we also performed our experiments after adding random noise to the future covariates.

## 2. METHODOLOGY & MODELING

In this section, we present the approach used. We introduce data acquisition, data processing, the HEC-RAS models, and the deep learning (DL) models. Both single and combinational DL models are explored in this paper. HEC-RAS is a widely used simulation software to model the hydraulics of water flow through natural rivers and other channels and is well-suited to our study domain – the coastal watershed in South Florida, for which we obtained a HEC-RAS model. Thus, HEC-RAS became our representative physics-based model to be used as a benchmark for comparisons.

### 2.1 Study area and Dataset

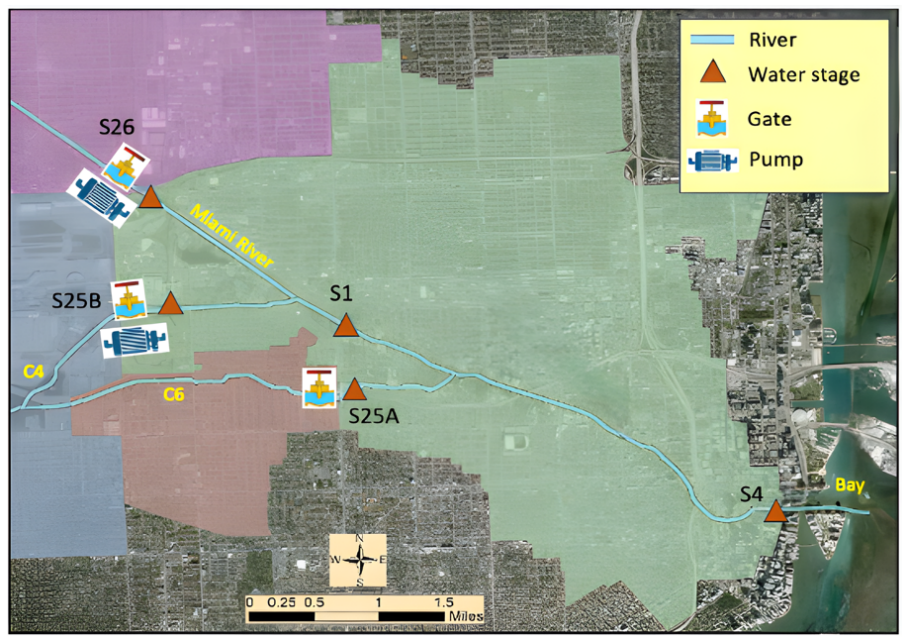

*Figure 1. Map of the study area, showing the shape of the downstream of Miami River, location of the hydrological stations, and waters stage measuring stations. The legend represents the elevation of the Datum NAVD 88.*

#### 2.1.1 Study area

South Florida is vulnerable to hurricane events with extremely heavy precipitation (Azzi et al., 2020). The Miami River is one of the largest rivers in South Florida emptying into the Biscayne Bay. The city of Miami with a sizable population and many commercial enterprises is located along the river. A schematic of the South Florida region is shown in Figure 1. The study area consists of 5.6 miles of the last part of the Miami River, its two tributaries: C4 (upper) and C6 (lower), and three water stations, S25A, S25B, and S26 located 1.13 miles, 0.94, and 5.61 miles upstream away from mainstream junctions of C6 and C4, respectively, and hydraulic structures (i.e., spillway gates, pumps) used to manage the flows. Pumps can allow additional discharge capacity during high tide or storm surge events. Water station S26 is a water station with a gated spillway and a set of pumps. Station S25B has two spillway gates that work independently and a set of pumps. The gate and pump flow measurements are the total flow through the gate and pump, respectively. Station S25A has a culvert structure (with a control gate), which is recorded. Stations S1 and S4 are water stations used to record and monitor the water stage. The mouth of the river is located on the Biscayne Bay, which demonstrates the downstream water stage is strongly influenced by the tides in Biscayne Bay. The hydraulic structures avoid seawater flowing back from the Biscayne Bay and thus have a significant impact on the river system's state.

*Table 1. Description of data feature and usage*

| Description | Abbreviation | Unit | Target variables |
|---|---|---|---|
| S26 gate flow | Flow_S26 | cubic feet / second | |
| S26 pump flow | Pump_S26 | cubic feet / second | |
| S26 tailwater stage | TWS_S26 | cubic feet / second | target |
| S25A gate flow | Flow_S25A | cubic feet / second | |
| S25A tailwater stage | TWS_25A | feet | target |
| S25B total gate flow | Flow_S25B | cubic feet / second | |
| S25B pump flow | Pump_S25B | cubic feet / second | |
| S25B tailwater stage | TWS_S25B | feet | target |
| S1 water stage | WS_S1 | feet | target |
| S4 water stage (tide info) | WS_S4 | feet | |
| Mean gridded precipitation intensity | Grid_Rainfall | inches / hour | |

### 2.1.2 Data acquisition and description

The river data used in this paper was collected hourly over a period of 11 years (Jan 1, 2010 to Dec 31, 2020) at five water stations (S1, S4, S25A, S25B, and S26 shown in Figure 1) and retrieved from DBHYDRO, the South Florida Water Management District's (SFWMD) database. The rainfall data is pixel-based (grided) radar precipitation data retrieved from the Next Generation Weather Radar (NEXRAD) system data archived by SFWMD. The precipitation data was collected every hour using a grid of sensors and averaged over all the grid cells in the domain. Other features, the flows at upstream locations (S25A, S25B, S26) and the water stage at downstream (S4), were additional covariates that can influence the prediction of the target variables (i.e., water stage at S1 and tail water stage at S25A, S25B, and S26). The data features and usage are described in Table 1, and all these inputs are used for both HEC-RAS and DL models.

### 2.1.3 Data preprocessing for Deep Learning Models

Unlike for the HEC-RAS model, data for the DL models need to be preprocessed beforehand. The data were normalized to avoid gradient values from exploding or vanishing (i.e., becoming infinity or zero). Min-Max normalization technique (Patro et al., 2015) was used as shown in Eq. (1):

$$x' = \frac{x - x_{min}}{x_{max} - x_{min}} \quad (1)$$

where the minimum value was set to 0, the maximum to 1, and all other values were linearly scaled as shown in (1). The dataset was divided into 80% training (Jan 1, 2010 to Aug 7, 2018) and 20% test set (Aug 8, 2018 to Dec 31, 2020).

## 2.2 HEC-RAS Model and Other Baseline Models

HEC-RAS was used to simulate a benchmark physics-based model. In this work, a 1D-2D coupled HEC-RAS model was used to calculate the overland flow more accurately. Four boundary conditions were required to run the HEC-RAS model: three upstream conditions (at S26, S25A, and S25B) and one downstream condition (at S4). All three upstream conditions were set as flow hydrographs. The upstream flow hydrograph was set to the gate flow value, with an added pump flow value, if there was a pump station. In HEC-RAS, the average of the gridded precipitation data over all grids in the domain was added to the boundary conditions at station S26. We have the water stage at S4 as downstream boundary condition for HEC-RAS. Since the HEC-RAS model is a numerical solver for shallow water equations, the input data (boundary conditions) of HEC-RAS had the same period ($t + 1$ to $t + k$) as the output data, and there was no lag time data needed in the model. A 2-year long simulation (from Jan 1, 2019, 00:00 to Dec 31, 2020, 23:00) was completed using HEC-RAS and was used to compare with the prediction results of DL models. The computation time step size was set at 1 minute to keep the computation stable, and the output interval was set the same as for the DL computations (i.e., 1 hour). The HEC-RAS modeling framework is provided in Appendix A.

We implemented the basic linear regression (LR) model to use as a baseline statistical approach. We also chose XGBoost as a baseline machine learning (ML) model. XGBoost was among the best performing ML model in prior experiments with this data set (data not shown).

## 2.3 Deep Learning Models

Four DL model architectures were used in this paper, including Recurrent Neural Networks (RNN), Long Short-Term Memory networks (LSTM), Multilayer Perceptron (MLP), and Convolutional Neural Networks (CNN). A DL model combining RNN and CNN in series was also developed to achieve better performance. The detailed architecture and mathematical representation of each model are outlined in Appendix B. The overall modeling framework for the DL models is shown in Figure 2. The significant difference between the DL models and HEC-RAS is that the DL methods depend on past information (e.g., looking back $w$ timesteps of all the variables), future predictable information (e.g., $k$ timesteps of rainfall data in the future), and human-controlled information (e.g., flow rate scheduled through the gate or pump for the $k$ timesteps), while HEC-RAS only requires the last two. In other words, HEC-RAS only needs the boundary conditions for the prediction period, while DL models were designed to use both the (future) information from the desired period as well as the data from the recent past.

In Figure 2, we assume that the current time is represented by time $t$, and $w$ and $k$ are the lengths of the history considered and the size of forecast lead time, respectively. WS represents the water stages at S1, S25A tailwater, S25B tailwater, and S26 tailwater. Gate flow refers to the amount of water flowing through the gate and Pump flow represents the amount of pumped water, both representing a time series. Tide is the water stage at S4, i.e., the sea stage information. For each location, we seek to predict the water stage from time $t + 1$ to $t + k$. Therefore, all covariates (rain, tide, gate flow, pump flow) from $t + 1$ to $t + k$ were considered using predicted or scheduled information. Rain and tide forecasts are available from the weather forecasting agency and oceanographic agency, while gate flow and pump flow are assumed to have been provided by the water management agency. DL models are also provided with historical data (rain, gate flow, pump flow, water stage) from time $t-w$ to $t$, with the assumption that recent past information informs us of the "state of the system". Mathematically, the modeling process is:

$$WS^i_{[t+1,t+k]} = f(WS_{[t-w,t]}, Tide_{[t-w,t+k]}, Rain_{[t-w,t+k]}, GF_{[t-w,t+k]}, PF_{[t-w,t+k]}), \qquad (2)$$

where $WS^i_{[t+1,t+k]}$ is water stage at the $i^{th}$ station from time $t+1$ to $t+k$, $i \in \{S1, S25A, S25B, S26\}$; $WS_{[t-w,t]}$ is a vector representing water stages for all four locations from time $t-w$ to $t$; $Tide_{[t-w,t+k]}$ is water stage at $S4$ from time $t-w$ to $t+k$; $Rain_{[t-w,t+k]}$ is the mean gridded rainfall for the study domain from time $t-w$ to $t+k$; $GF_{[t-w,t+k]}, PF_{[t-w,t+k]}$ represent water flow through gate and pump respectively for all four stations from time $t-w$ to $t+k$. We set the length of past window (also called look-back window (Nie et al., 2023), rolling window (Li et al., 2014), sliding window (Shi et al., 2023)) $w = 72$ hours and the size of forecast lead time $k = 24$ hours in our work.

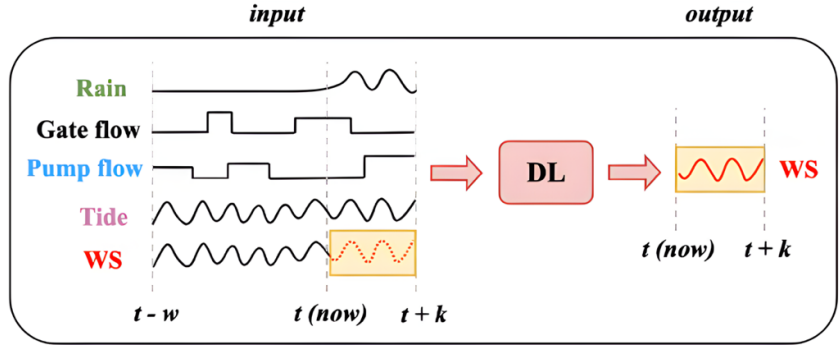

*Figure 2. Overall framework of the DL models. Vertical dashed lines indicate time points of interest. Input data is shown as a set of time series to the left of the schematic (except water stage from time $t+1$ to $t+k$), while the output to predict is shaded yellow.*

### 2.3.1 Multilayer Perceptron (MLP)

Multilayer Perceptron (MLP) is part of a family of fully connected feedforward neural networks (Naskath et al., 2023). MLP usually consists of at least three layers: one input layer, one or more hidden layers, and one output layer. The input layer is used to load input features to be processed. Hidden layers and the output layer are used for intermediate non-linear computations and processing the predicted output, respectively. By stacking multiple layers of neurons, where each layer applies a non-linear activation function, the MLP can learn complex, non-linear mappings from inputs to outputs (Del Campo et al., 2021). The composition of these non-linear functions through the network allows it to approximate highly intricate functions. However, since MLPs are fully connected neural networks, they are prone to overfit the training data, leading to poor generalization on new, unseen data (Zelený et al., 2023). Therefore, *Regularization* and *Dropout* techniques are generally used to mitigate overfitting by simplifying the model structure.

### 2.3.2 Recurrent Neural Networks (RNNs)

Unlike traditional feedforward neural networks, Recurrent Neural Networks (RNNs) have connections that form directed cycles, allowing them to maintain a 'memory' of previous inputs (Rezk et al., 2020). This is particularly useful for tasks where the context from previous inputs is important. They are well suited to recognize patterns in sequences of data, such as time series, text, speech, or video. The temporal dependencies can be learned by recurrently training and updating the transitions of an internal (hidden) state from the last timestep to the current timestep. Nevertheless, RNNs have the following limitations. Since RNNs learn knowledge iteratively over time steps, training RNNs can be challenging due to the vanishing or exploding gradient problem, especially for long sequences (Kag et al., 2020). Specifically, the gradients at the last time steps are difficult to backpropagate to the time steps at the beginning. This phenomenon is also called "memory forgetting" (Ghojogh et al., 2023), caused by the inherent characteristic of RNNs - sequential modeling.

### 2.3.3 Long Short-term Memory (LSTM)

RNNs are susceptive to forget the knowledge from the past due to vanishing gradients, LSTM networks were proposed as a variant of RNNs (Shi et al, 2022; Rithani et al., 2022) to mitigate this problem. At time point $t$, an LSTM includes a cell state, $C_t$, which runs straight along the entire chain with only some linear interactions. LSTM has hidden states, $S_t$, representing the information from the past timesteps and the information that is filtered/updated with the new input $x_t$ by the following gate operations. Generally, LSTM can learn slightly longer time dependencies by incorporating forget gates, input gates, additional gates, and output gates to control the inflow, filtering, and outflow of past and current information. The above gates can help the models to alleviate the vanishing gradient problem to some extent. However, LSTMs are more complex than traditional RNNs due to the multiple gates and state variables, leading to longer training times (Yu et al., 2019).

### 2.3.4 Convolutional Neural Networks (CNNs)

Convolutional Neural Networks (CNNs), traditionally used for image and video processing, have also been proven effective for time series forecasting. When applied to time series data, CNNs can capture temporal patterns and dependencies by treating the time series as a one-dimensional sequence. Convolutional layers apply filters (kernels) that slide over the input sequence to detect patterns (Borovykh et al., 2018). Each filter is a small, learnable array that extracts features from the time series. Each convolutional layer is usually followed by a maximum pooling layer, which is used to extract the most meaningful values of each convolutional computation. When multiple time series (features) are provided, the input can be represented as a two-dimensional array (time steps × features). This allows the CNN models to capture and learn the relationships across different variables (Keren & Schuller, 2016).

In summary, CNNs excel at capturing local patterns in data since each filter has a local receptive field, meaning it looks at a small portion of the input at a time. More interestingly, CNNs apply filters uniformly across the input, so their ability to detect patterns is not affected by the length of the sequence (Alzubaidi et al., 2021). They do not rely on propagating information through a hidden state, so the prediction quality remains consistent regardless of sequence length. On the other hand, CNNs may not capture sequential dependencies as effectively as RNNs due to the limited size of local receptive field. More details of CNNs can be found in (Keren & Schuller, 2016, Borovykh et al, 2017).

### 2.3.5 Combination of RNN & CNN: RCNN

To leverage the strengths of both RNNs and CNNs while mitigating their weaknesses, an approach is to use an RNN first, followed by a CNN. This configuration can be particularly advantageous for time series forecasting and other sequential tasks. Initially, an RNN can process the sequential data, capturing temporal patterns and sequential dependencies. The RNN's hidden states at each time step encapsulate the temporal dynamics of the sequence. These hidden states, which now contain rich temporal information, can then be treated as a new sequence of features. Next, a CNN can be applied to this sequence of RNN-generated features. The convolutional layers in the CNN can then detect local patterns and spatial hierarchies within the sequence of RNN features. This combination allows the CNN to extract meaningful patterns from the temporally aware features produced by the RNN, enhancing the model's ability to recognize both short-term and long-term dependencies.

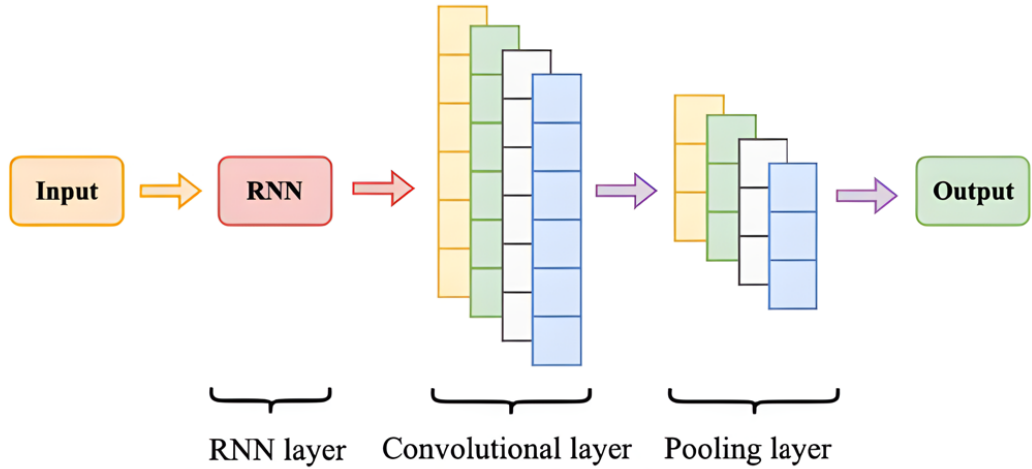

*Figure 3. Architecture of the Recurrent Convolutional Neural Networks*

## 3. RESULTS

For DL models, we used all the measured variables from the past 168 timesteps (7 days) and all the predictable and determinate covariates (precipitation and gate releases) from the future 24 timesteps (1 day) as input data, with the intent to predict target variables (water stage at S1, S25A, S25B, S26) for the future 24 timesteps (1 day). After training DL models with the training set, we applied these models to the test set (2019-2020) and compared the predicted water stages with the simulated water stages of HEC-RAS and the observed water stages from the measuring stations. In the following discussion, we represent the $k^{th}$ predicted timestep in the future with $t + k$ hour prediction. For example, $t + 24$ hour prediction refers to the predicted timestep 24 hours in the future.

### 3.1 Performance analysis

### 3.1.1 Performance analysis on prediction timesteps

DL models were set to predict the target variables from $t + 1$ hour to $t + k$ hours. In this subsection, we present the prediction results for the forecast lead times $t + 1, t + 12, t + 18, t + 24$ hours. The experimental results are reported using "perfect foresight" future covariates (tide and rainfall). We have included linear regression (LR) and a machine learning model (XGBoost) as baselines. While the physics-based model, HEC-RAS, has also been included, we note that it does not provide "predictions" in the true sense of the term with arbitrary forecast lead times. Instead, it provides simulations using "perfect foresight" covariates at that future time. Consequently, the error in simulation for HEC-RAS is independent of the lead time. In contrast, all machine learning models can, in theory, predict with arbitrary lead times with or without the availability of future covariate estimations. Metrics including mean absolute error (MAE), root mean square error (RMSE), Nash–Sutcliffe model efficiency coefficient (NSE), and Kling–Gupta efficiency (KGE) were used to measure the difference between observed and predicted data. NSE and KGE are goodness-of-fit indicators widely used in hydrologic sciences for comparison between simulations to observations, while MAE and RMSE are widely used in the machine learning community.

$$MAE = \frac{\sum_{i=1}^{N} |y_i - \hat{y}_i|}{N}, \qquad (3)$$

$$RMSE = \sqrt{\frac{\sum_{i=1}^{N} (y_i - \hat{y}_i)^2}{N}}, \qquad (4)$$

$$NSE = 1 - \frac{\sum_{i=1}^{N} (y_i - \hat{y}_i)^2}{\sum_{i=1}^{N} (y_i - \bar{y}_i)^2}, \qquad (5)$$

$$KGE = 1 - \sqrt{(r-1)^2 + (\alpha-1)^2 + (\beta-1)^2}, \qquad (6)$$

where $y_i$, $\hat{y}_i$ represent the observed data and model outputs, $\bar{y}_i$ is the mean of observed water levels, and $N$ denotes the number of samples. In Eq. (6), $r$ is the Pearson correlation coefficient, $\alpha$ is a term representing the variability of prediction errors, and $\beta$ is a bias term.

Table 2 shows the performance as a function of forecast lead time. The initial observation reveals that deep learning (DL) models outperform linear regression and traditional machine learning models. This underscores their superior capability to learn and capture complex non-linear relationships within the data. Among DL models, RNN and LSTM models were more sensitive to the forecast lead time, performing well for short-term predictions (1-hr and 8-hr), but suffered larger errors for longer-time predictions. In contrast, MLP and CNN were relatively insensitive to forecast lead time, displaying stable predictions with lower variance even for longer predictions. The combined model (RCNN) produced the least MAE and in four out of five cases for the RMSE measure. We surmise that RCNNs can simultaneously learn the cross-time and cross-variable relationships from data.

In our experiments above, we used the actual future observed values for the future covariates, but they are estimated with some uncertainties in real scenarios. To mimic the uncertainty in future estimates, we follow prior work by adding random Gaussian noise as an uncertainty proxy (Tabas & Samadi, 2022; Iliopoulou & Koutsoyiannis, 2020). Therefore, we also performed experiments after adding 20% random noise to these covariates over the entire future $k$ time steps during the test phase. The magnitude of the noise signal utilized in our experiments was suggested by the work on quantitative precipitation forecast by SFWMD (Seo et al., 2012). We observed that the fluctuation of prediction errors by using the DL methods (see Table 3) is relatively small (< 4%). Even when 40% random noise values were added as a proxy to the forecast estimates, the errors roughly doubled (data not shown). However, the error percentages remained low yet, suggesting that the DL models capture the underlying relationships in the data rather than relying on specific, possibly noisy, patterns in the training data.

*Table 2. Performance comparison of DL models at all locations (S1, 25A, S25B, S26) for different forecast lead times. The lowest prediction errors (MAE, RMSE) and highest efficiencies (NSE, KGE) are in blue and red, respectively.*

| Metrics | Models | 1-hr ahead | 8-hr ahead | 16-hr ahead | 24-hr ahead | Entire 24 hrs |
|---|---|---|---|---|---|---|
| MAE (ft) | HEC-RAS | 0.186 | 0.186 | 0.186 | 0.186 | 0.186 |
| | LR | 0.275 | 0.199 | 0.261 | 0.288 | 0.296 |
| | XGBoost | 0.184 | 0.187 | 0.188 | 0.192 | 0.187 |
| | MLP | 0.161 | 0.167 | 0.148 | 0.157 | 0.151 |
| | RNN | 0.075 | 0.093 | 0.145 | 0.148 | 0.120 |
| | LSTM | 0.098 | 0.147 | 0.165 | 0.152 | 0.148 |
| | CNN | 0.102 | 0.120 | 0.118 | 0.124 | 0.118 |
| | RCNN | 0.097 | 0.094 | 0.099 | 0.106 | 0.098 |
| RMSE (ft) | HEC-RAS | 0.238 | 0.238 | 0.238 | 0.238 | 0.238 |
| | LR | 0.256 | 0.318 | 0.317 | 0.347 | 0.332 |
| | XGBoost | 0.270 | 0.273 | 0.274 | 0.284 | 0.274 |
| | MLP | 0.206 | 0.205 | 0.183 | 0.197 | 0.191 |
| | RNN | 0.098 | 0.126 | 0.185 | 0.192 | 0.158 |
| | LSTM | 0.124 | 0.198 | 0.218 | 0.202 | 0.200 |
| | CNN | 0.130 | 0.153 | 0.151 | 0.157 | 0.150 |
| | RCNN | 0.118 | 0.122 | 0.130 | 0.141 | 0.129 |
| NSE | HEC-RAS | 0.932 | 0.932 | 0.932 | 0.932 | 0.932 |
| | LR | 0.920 | 0.877 | 0.878 | 0.854 | 0.866 |
| | XGBoost | 0.912 | 0.910 | 0.910 | 0.903 | 0.910 |
| | MLP | 0.948 | 0.949 | 0.959 | 0.953 | 0.956 |
| | RNN | 0.988 | 0.980 | 0.958 | 0.955 | 0.969 |
| | LSTM | 0.981 | 0.953 | 0.942 | 0.951 | 0.951 |
| | CNN | 0.979 | 0.971 | 0.972 | 0.970 | 0.972 |
| | RCNN | 0.980 | 0.982 | 0.979 | 0.976 | 0.980 |
| KGE | HEC-RAS | 0.950 | 0.950 | 0.950 | 0.950 | 0.950 |
| | LR | 0.735 | 0.841 | 0.770 | 0.710 | 0.731 |
| | XGBoost | 0.830 | 0.822 | 0.820 | 0.816 | 0.821 |
| | MLP | 0.882 | 0.880 | 0.893 | 0.888 | 0.887 |
| | RNN | 0.958 | 0.951 | 0.890 | 0.907 | 0.920 |
| | LSTM | 0.922 | 0.895 | 0.885 | 0.910 | 0.895 |
| | CNN | 0.924 | 0.898 | 0.913 | 0.907 | 0.905 |
| | RCNN | 0.933 | 0.936 | 0.931 | 0.936 | 0.936 |

To analyze the distribution of prediction errors, violin plots were used to visualize the prediction errors for 24-hours in the future (see Figure 4). Violin plots display the density distribution along with the median value, the peak value, the 25$^{th}$ percentile, and the 75$^{th}$ percentile (see the top and bottom of the small black box inside). The larger the highest density of the violin plot, the greater are the errors distributed in that corresponding area. We also observe that the majority of the prediction errors for the RCNN model are centered around 0 feet, predominantly within the narrow range of -0.3 to 0.3 feet. Compared to other

DL models, the RCNN model exhibited lower frequency and magnitude for extreme errors. This trend of lower mean and variance in prediction errors remains consistent even as the forecast lead times increase, demonstrating the RCNN model's superior accuracy and stability over time. The superior performance of RCNN was also confirmed by the p-values for the non-parametric Wilcoxon test used to compare RCNN with the other four DL models. All comparisons between RCNN and other models were statistically significant with p values < 0.01 (see Appendix D1).

*Table 3. MAE Changes of the DL models of all locations (S1, 25A, S25B, S26) for different forecast lead times after adding random noise to the observed precipitation.*

| Models | 1-hr ahead | 8-hr ahead | 16-hr ahead | 24-hr ahead | Entire 24 hrs |
|---|---|---|---|---|---|
| MLP | ↑0.047% | ↑0.097% | ↑0.027% | ↓0.005% | ↑0.050% |
| RNN | ↑0.001% | ↓0.004% | ↓0.015% | ↓0.014% | ↓0.001% |
| LSTM | ↓0.005% | ↑0.000% | ↑0.001% | ↑0.003% | ↓0.002% |
| CNN | ↑1.222% | ↑0.535% | ↑3.320% | ↑2.139% | ↑1.615% |
| RCNN | ↑0.116% | ↑0.118% | ↑0.128% | ↑0.061% | ↑0.105% |

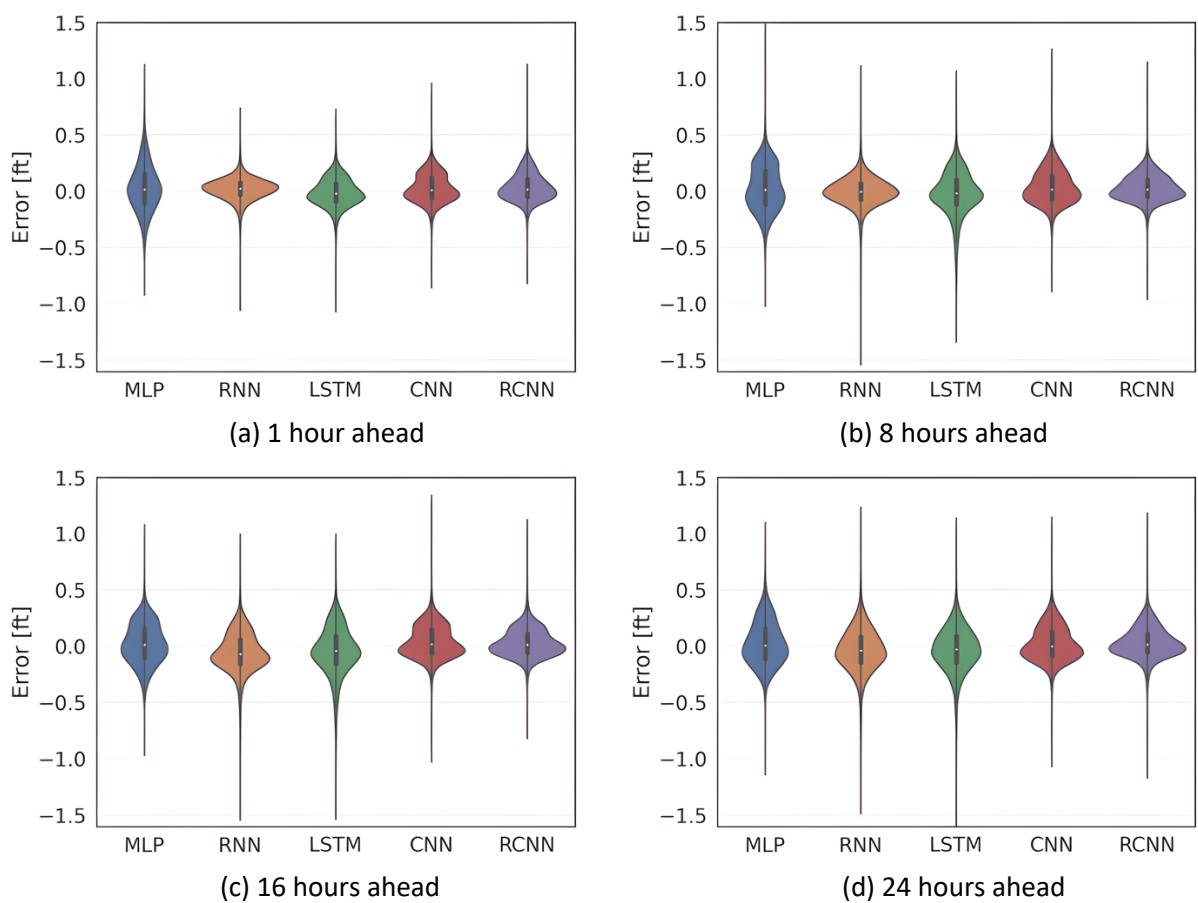

*Figure 4. Violin plots showing density distributions of actual errors for the five DL methods at all locations with forecast lead time: (a) 1 hour; (b) 12 hours; (c) 18 hours; and (d) 24 hours. We zoomed in by magnifying the y-axis of the violin plots for clear comparison (see Appendix C2.1). Note that the prediction errors in violin plots are reported based on the perfect foresight of future covariates (tide and rainfall).*

Additionally, to analyze the extreme prediction errors quantitively, we reported those extreme errors that are beyond the range (-0.5, 0.5) feet in Table 4. We note that 99 percent of the absolute prediction errors of the DL models within 0.5 feet (Note that our programs allow the threshold to be changed flexibly). It shows the RCNN model has the lowest percentage (0.803%) of error extremes. More interestingly, RNN and LSTM models generate consistently higher magnitude of error extremes with the increase in forecast lead time while MLP, CNN, and RCNN perform relatively stable.

Table 4. Percentage of prediction errors beyond the range (-0.5, 0.5) feet. The lowest percentages are in blue.

| Models | 1-hr ahead | 8-hr ahead | 16-hr ahead | 24-hr ahead | Entire 24 hrs |
|---|---|---|---|---|---|
| MLP | 1.684 % | 0.880 % | 0.358 % | 0.990 % | 0.753 % |
| RNN | 0.073 % | 0.464 % | 1.316 % | 1.481 % | 0.777 % |
| LSTM | 0.110 % | 2.499 % | 3.125 % | 2.237 % | 2.360 % |
| CNN | 0.086 % | 0.237 % | 0.297 % | 0.291 % | 0.244 % |
| RCNN | 0.127 % | 0.194 % | 0.250 % | 0.283 % | 0.224 % |

### 3.1.2 Comparing the performance of models at different locations

We present MAE results to study the model performance across different locations. In Table 5, we observed that the DL models display smaller MAE values than HEC-RAS for all locations. RCNN model outperforms all other models for 3 of 4 locations (S1, S25A, S25B). On the other hand, location S26 has the highest errors for all DL models compared to other locations.

Table 5. Mean absolute error (MAE) of the predicted results for 24 hours at four different locations for the test set.

| Models | S1 | S25A | S25B | S26 |
|---|---|---|---|---|
| MLP | 0.160 | 0.106 | 0.105 | 0.258 |
| RNN | 0.135 | 0.139 | 0.150 | 0.170 |
| LSTM | 0.146 | 0.136 | 0.160 | 0.167 |
| CNN | 0.107 | 0.093 | 0.090 | 0.206 |
| RCNN | 0.072 | 0.085 | 0.082 | 0.182 |
| HEC-RAS | 0.176 | 0.182 | 0.198 | 0.184 |

Table 6. Percentage of prediction errors beyond the range (-0.5, 0.5) feet at four different locations.

| Models | S1 | S25A | S25B | S26 |
|---|---|---|---|---|
| MLP | 0.208 % | 0.229 % | 0.322 % | 2.847 % |
| RNN | 0.961 % | 1.455 % | 1.704 % | 1.273 % |
| LSTM | 2.364 % | 1.320 % | 2.883 % | 1.579 % |
| CNN | 0.109 % | 0.208 % | 0.218 % | 0.525 % |
| RCNN | 0.094 % | 0.338 % | 0.327 % | 0.374 % |
| HEC-RAS | 2.599 % | 3.249 % | 4.417 % | 3.585 % |

Figure 5 shows that the DL models perform with lower magnitude of error extremes at location S1 (results for locations S25A, S25B, and S26 are in Appendix C1). Based on the distribution of prediction errors, we observed that the medians of DL models are comparable with that of HEC-RAS. Furthermore, most prediction errors of DL models are in the (-0.25, 0.25) feet range when compared to HEC-RAS (-0.4, 0.4) feet, suggesting a lower variance for the DL models. In Table 6, we computed the percentage of extreme errors beyond the range (-0.5, 0.5) feet.

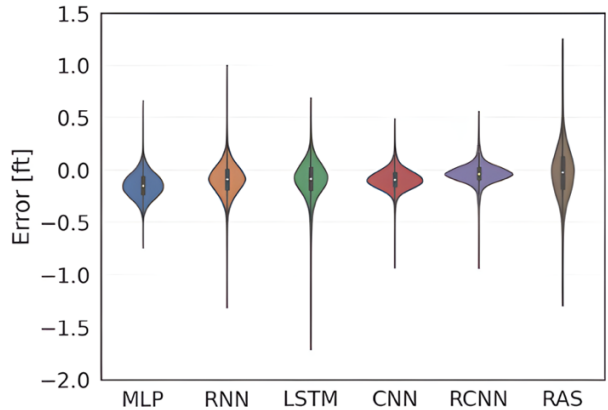

*Figure 5. Violin plots of error distributions for 24-hours-ahead forecasts at locations, S1. All comparisons with HEC-RAS (RAS) were statistically significant at the 0.01 level. Figures in the Appendix C2.2 shows difference among these same violin plots but enlarged in the middle to see the differences more clearly.*

### 3.2 Result visualization

In this section, we present visualizations of the observed and predicted water stages at station S1 In this section, we present visualizations of the observed and predicted water stages at location S1, which is centrally located near Miami International Airport. The visualization for other stations has been provided in Appendix C4. We chose two short periods: one without a storm event (April 1, 2019) and another with a storm event (August 1-3, 2020).

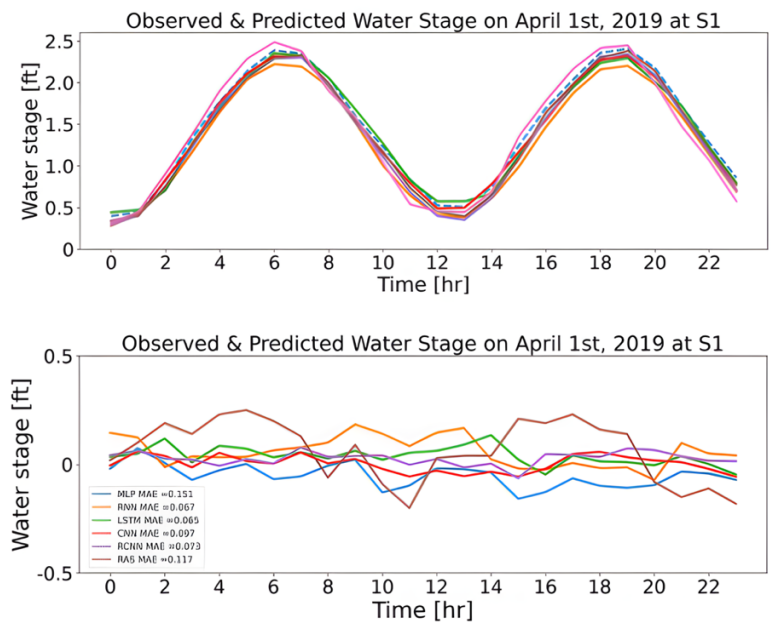

*Figure 6. Observed & Predicted Water Stage at station S1 on April 1, 2019.*

### 3.2.1 Normal period without a storm event

Figure 6 shows the observed water stages, predicted water stages (using the DL models), and the simulated water stages (using HEC-RAS) at location S1 on April 1, 2019, which had no storm events. The corresponding prediction errors have been also presented below. Both DL models and HEC-RAS models showed good agreement with the observed data. Moreover, DL models show the lower prediction errors, indicating their powerful capacity in water stage prediction.

### 3.2.2 Period with an extreme storm event

Heavy storm events have great impacts on water stages in the river. Accurately predicting the water stages is critical during such periods. In this work, we used around 9 years of historical data to train the DL models. The training dataset included extreme events like Hurricane Irma (2017), tropical storm Andrea (2013), Hurricane Sandy (2012), and tropical storm Isaac (2012). Tropical storm Isaias (2020) was used to test the accuracy of the models. Tropical storm Isaias reached the greater Miami region nearby around August 1-3, 2020. Thus, we chose this period for the analysis as well.

The predictions of 24-hour ahead during tropical storm Isaias and the corresponding prediction errors are shown in Figure 7. Compared to the normal conditions, the water stages during tropical storm Isaias are higher. Meanwhile, the MAE values in this set of experiments are slightly higher compared to the experiments in Section 3.2.1. However, all DL models still outperformed the HEC-RAS, with RCNN continuing to rank the first among the DL models. Visualization for locations S25A, S25B, and S26 are provided in Appendix C4.

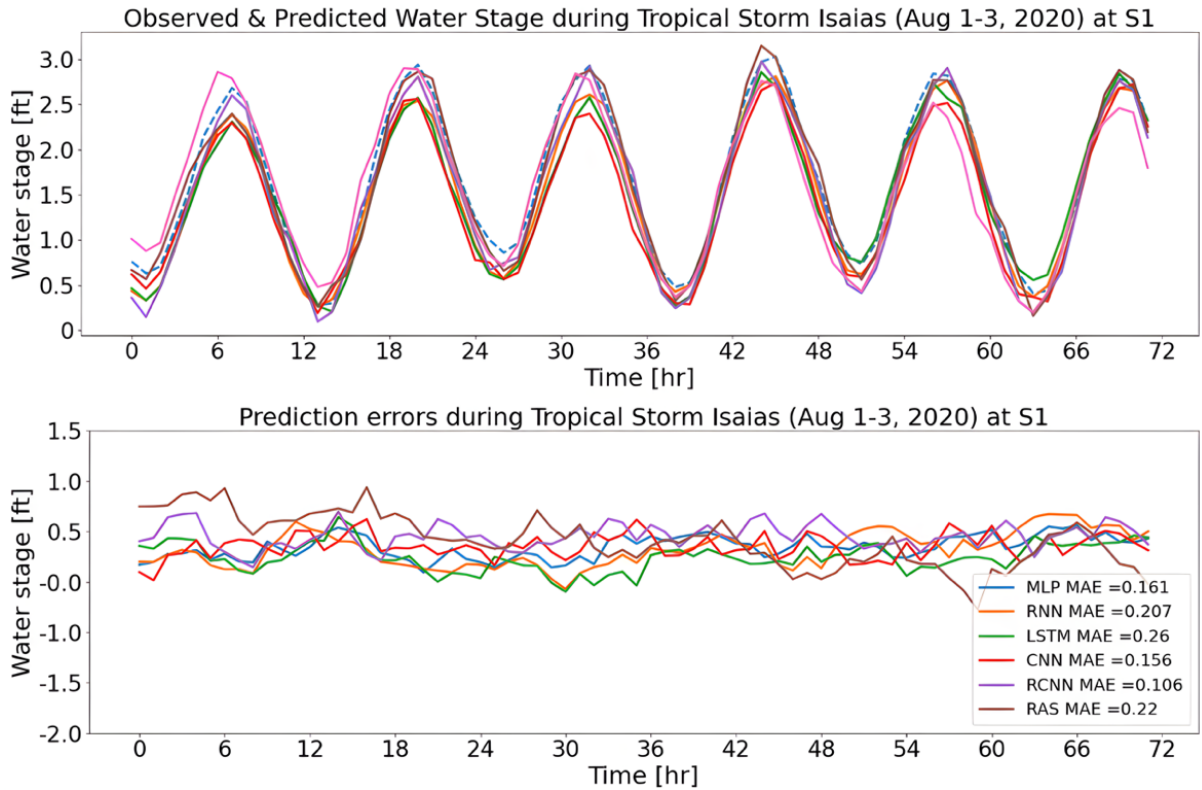

*Figure 7. Observed and Predicted Water Stages at station S1 during Tropical Storm Isaias (2020).*

**3.3 Computation time**

Rapidly predicting the water stage is crucial, if these tools are to be used during real-time operations and with different hypothetical boundary conditions (i.e., different gate flow, pump flow, rainfall intensity, tide). Table 7 compares the run times for the DL models in their test phase with HEC-RAS. The training phase is a one-time cost and needs to be included here.

*Table 7. Training and test time for all models.*

|  | HEC-RAS | MLP | RNN | LSTM | CNN | RCNN |
|---|---|---|---|---|---|---|
| Training time | - | 27 mins | 267 mins | 1033 mins | 167 mins | 667 mins |
| Test time | 45 min | 0.34 s | 1.81 s | 4.81 s | 0.62 s | 2.41 s |

**4. DISCUSSION & CONCLUSIONS**

Several DL models were developed and tested to predict water stages at specific locations in the river. Table 7 shows the execution time of the DL models is much lower than the HEC-RAS model, thus bringing us closer to building real-time prediction tools for more complex tasks. In terms of the model accuracy, all DL models in our paper show a good agreement with the observed data and their performance is better than that of HEC-RAS. In addition, inaccurate predictions can be more dangerous especially when water levels are high. Overestimation of water stages can cause false alarms and panic, while underestimation of water stages can result in unexpected flooding events. We set the 99th percentile of water stage (i.e., 0.5 feet) as a cutoff and analyzed the prediction errors of underestimation and overestimation. In future work, we aim to deal with the small proportion of error extremes since floods could be occurring even during a short time.

The key conclusions have been listed as follows. Our deep learning (DL) surrogate models (MLP, RNN, CNN, LSTM, and RCNN) can predict water stages at locations of interest with higher accuracy and efficiency than the physics-based model, HEC-RAS. DL models show at least 500x speedup over HEC-RAS. More interestingly, we found that DL models such as RNN and LSTM were more sensitive to the forecast lead time, with higher prediction errors for greater forecast lead time while MLP, CNN, and RCNN models had a relatively stabler performance. The combinational RCNN model provides relatively better performance for both short and long-term prediction. Finally, the utilization of future covariates is essential for the predictions especially under extreme storm events. Our DL models indicate good robustness even when 20% random noise values were added to future covariates used in this work.

**REPRODUCIBILITY STATEMENT**

Data set and code with detailed instructions (readme file) are available in the GitHub repository: https://github.com/JimengShi/DL-WaLeF and Zenodo: https://zenodo.org/records/14715390.

**ACKNOWLEDGMENTS**

This work was partly supported by the National Science Foundation grants numbered 2203292 (to AL) and 2118329 (to GN via University of Illinois Urbana-Champaign).

Shi, J., Jain, M., & Narasimhan, G. (2022). Time Series Forecasting Using Various Deep Learning Models. International Journal of Computer and Systems Engineering, 16(6), 224-232.

Tabas, S. S., & Samadi, S. (2022). Variational Bayesian dropout with a Gaussian prior for recurrent neural networks application in rainfall–runoff modeling. Environmental Research Letters, 17(6), 065012.

Yin, Z., Zahedi, L., Leon, A.S., Amini, M.H. and Bian, L., A Machine Learning Framework for Overflow Prediction in Combined Sewer Systems. In World Environmental and Water Resources Congress 2022 (pp. 194-205).

Yu, Y., Si, X., Hu, C., & Zhang, J. (2019). A review of recurrent neural networks: LSTM cells and network architectures. *Neural computation*, *31*(7), 1235-1270.

Zelený, O., & Fryza, T. (2023, April). Multi-Branch Multi Layer Perceptron: A Solution for Precise Regression using Machine Learning. In 2023 33rd International Conference Radioelektronika (RADIOELEKTRONIKA) (pp. 1-5). IEEE.

Zhang, R., Zen, R., Xing, J., Arsa, D. M. S., Saha, A., & Bressan, S. (2020). Hydrological Process Surrogate Modelling and Simulation with Neural Networks. Advances in Knowledge Discovery and Data Mining, 12085, 449.

Zhao, G., Pang, B., Xu, Z., Peng, D., & Xu, L. (2019). Assessment of urban flood susceptibility using semi-supervised machine learning model. Science of the Total Environment, 659, 940-949.18

# Appendix

## A. HEC-RAS modeling

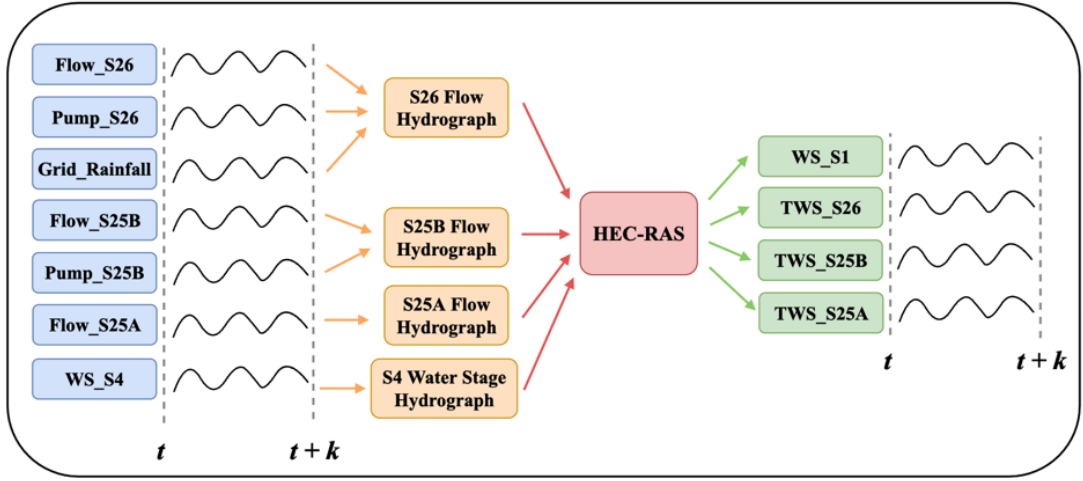

Figure 8. HEC-RAS modeling framework.

## B. Deep Learning Models

### B1. Multilayer Perception

Mathematically, it is described as follows:

$$\boldsymbol{h_1} = \sigma(W_1 \cdot \boldsymbol{x} + \boldsymbol{b_1});$$
$$\boldsymbol{h_2} = \sigma(W_2 \cdot \boldsymbol{h_1} + \boldsymbol{b_2});$$
$$……$$
$$\boldsymbol{h_l} = \sigma(W_l \cdot \boldsymbol{h_{l-1}} + \boldsymbol{b_l});$$
$$\boldsymbol{y} = \sigma(W_{l+1} \cdot \boldsymbol{h_l} + \boldsymbol{b_{l+1}}),$$

where $x$ is a flattened vector consisting of all inputs. The input layer is the $0^{th}$ layer, hidden layers $h_1$, $h_2$, …, $h_l$ are the next $l$ layers, and the output layer is the $(l+1)^{th}$ layer, $W_j \in \mathbb{R}^{n(j) \times n(j-1)}$ is the parameter matrix from the $(j-1)^{th}$ hidden layer to $j^{th}$ hidden layer, $n(j)$ is the number of neurons of $j^{th}$ layer; $b_j \in \mathbb{R}^{n(j)}$ is the bias vector of $j^{th}$ layer.

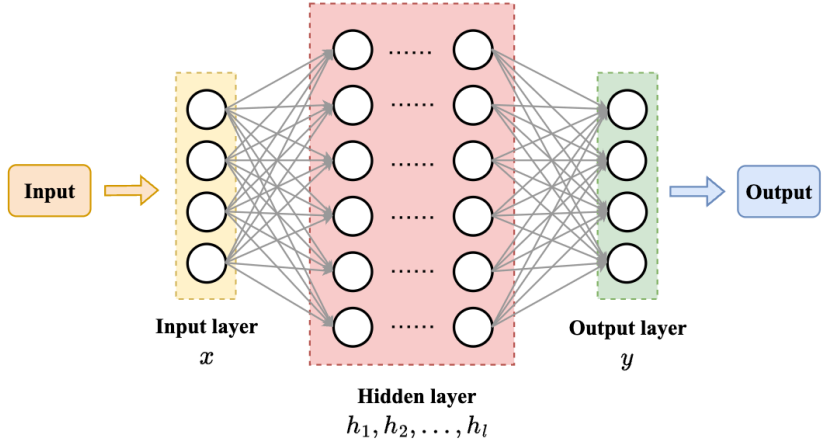

*Figure 9. Architecture of a Multilayer Perceptron*

### B2. Recurrent Neural Networks (RNNs)

Fig. 5 shows that RNNs can memorize what has already been processed before time $t$ and save the information in the hidden state $S_{t-1}$. RNNs combine current information, $x_t$, in state $S_t$, to generate output $y_t$. The comparison with the ground truth of $y_t$ allows RNNs to learn from previous iterations. The training process is determined by three parameter matrices $W_x$, $W_s$, $W_y$, and two bias vectors $b_s$, $b_y$. The simplified description of RNNs is that they propagate forward, predict output, and compute prediction errors. To minimize the errors, RNNs backpropagate the computed errors to update the gradients at each timestep and update weights for the next iteration. The computational process of each hidden state (unit or cell) is described in Fig. 6. Mathematically, it is given as follows:

$$S_t = tanh(W_{xs} \cdot (x_t \oplus S_{t-1}) + b_s) \text{ and}$$

$$y_t = \sigma(W_y \cdot S_t + b_y),$$

where $x_t \in \mathbb{R}^m$ is the input vector of $m$ input features at time $t$; $W_{xs} \in \mathbb{R}^{n \times (m+n)}$ and $W_y \in \mathbb{R}^{n \times n}$ are parameter matrices; $n$ is the number of neurons in the RNN layer; $b_s \in \mathbb{R}^N$ and $b_y \in \mathbb{R}^N$ are bias vectors for the internal state and output, respectively; $\sigma$ is the sigmoid activation function; $S_t$ is the internal (hidden) state; and $x_t \oplus S_{t-1}$ is the concatenation of vectors, $x_t$ and $S_{t-1}$.

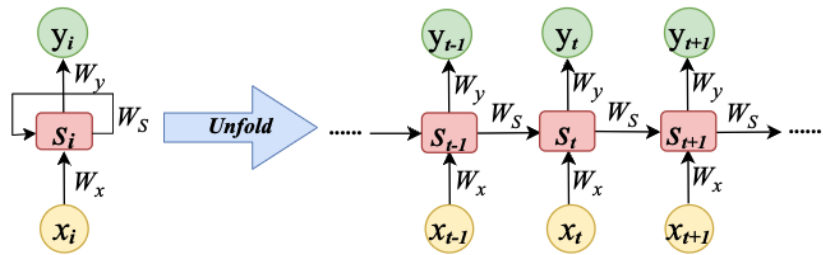

*Figure 10. Architecture of the Recurrent Neural Networks*

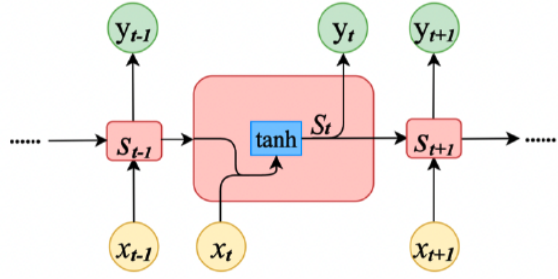

*Figure 11. Computation process of the Recurrent Neural Networks*

**B3. Long Short-term Memory (LSTM)**

The mathematical formulation is as follows:

$$f_t = \sigma(W_f \cdot (x_t \oplus S_{t-1}) + b_f);$$
$$i_t = \sigma(W_i \cdot (x_t \oplus S_{t-1}) + b_i);$$
$$\widetilde{C}_t = tanh(W_C \cdot (x_t \oplus S_{t-1}) + b_C);$$
$$C_t = f_t \cdot C_{t-1} + i_t \cdot \widetilde{C}_t;$$
$$O_t = \sigma(W_O \cdot (x_t \oplus S_{t-1}) + b_O);$$
$$S_t = tanh(C_t) \cdot O_t; \text{ and}$$
$$y_t = \sigma(W_y \cdot S_t + b_y),$$

where $x_t \in \mathbb{R}^m$ is the vector of $m$ input features at time $t$; $W_f, W_i, W_C, W_O \in \mathbb{R}^{n \times (m+n)}$ and $W_y \in \mathbb{R}^{n \times n}$ are parameter matrices; $n$ is the number of neurons in the LSTMs layer; $b_f, b_i, b_C, b_O, b_y \in \mathbb{R}^n$ are bias vectors; $\sigma$ is the sigmoid activation function; and $S_t$ is the internal (hidden) state. The functions $f_t, i_t, \widetilde{C}_t$, and $O_t$ are implemented by a forget gate, input gate, addition gate, and output gate, respectively.

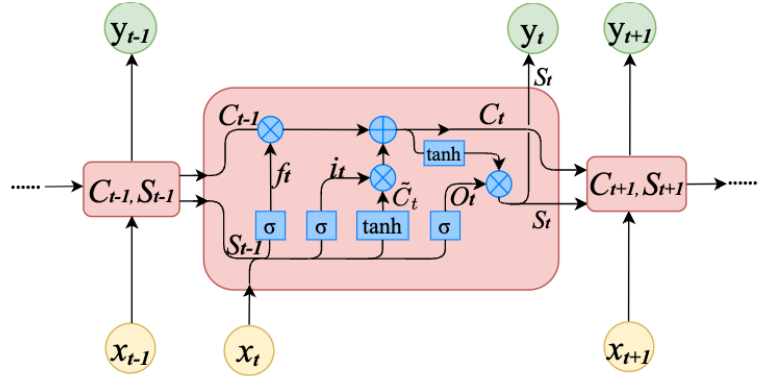

*Figure 12. Architecture of the Long Short-term Memory*

### B4. Convolutional Neural Networks (CNNs)

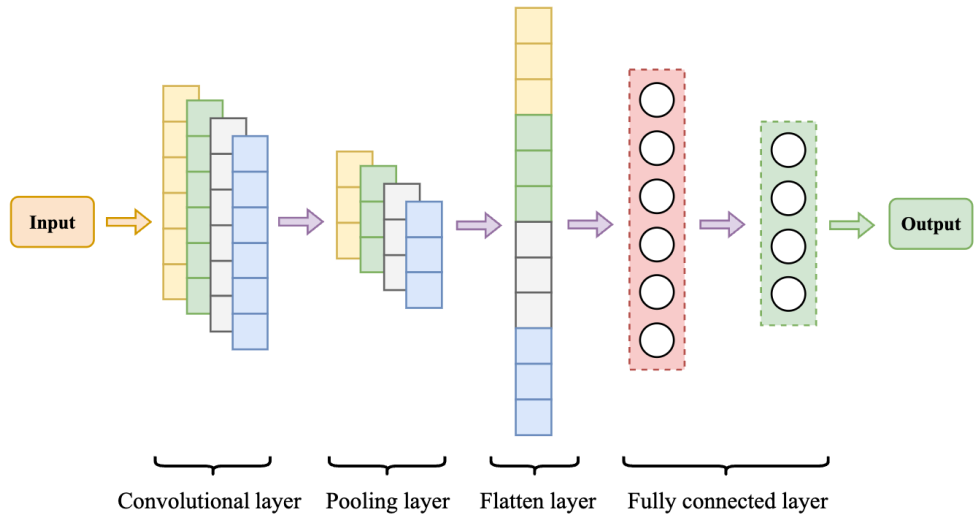

*Figure 13. Architecture of the Convolutional Neural Networks*

## C. Visualization

### C1. Violin plots at all locations

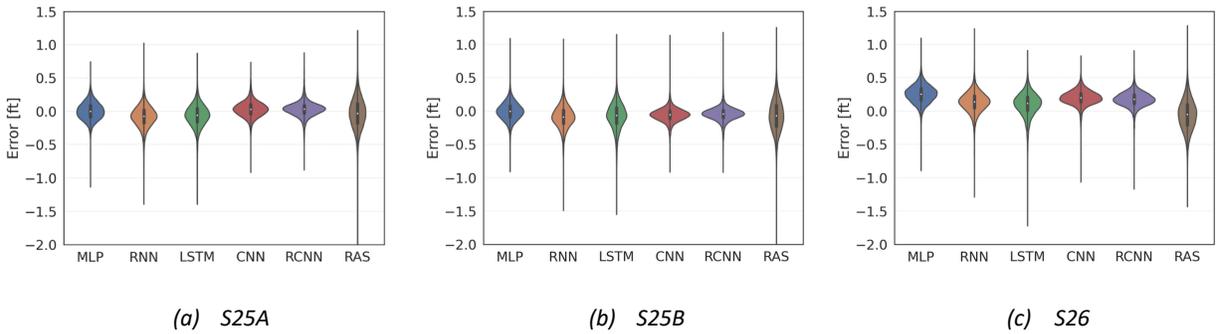

*(a)  S25A*  *(b)  S25B*  *(c)  S26*

*Figure 14. Violin plots of error distributions for 24-hours-ahead forecasts at locations S15A, S25B, and S26. All comparisons with HEC-RAS (RAS) were statistically significant at the 0.01 level. Appendix C2.2 shows difference among these same violin plots but enlarged in the middle to see the differences more clearly.*

### C2. Violin plots expanded

To observe the difference among these violin plots, we zoomed in Figures 5, 6 in the main paper by shrinking the range of y axis in Figures 15, 16, respectively.

*C2.1 Violin plots expanded over time*

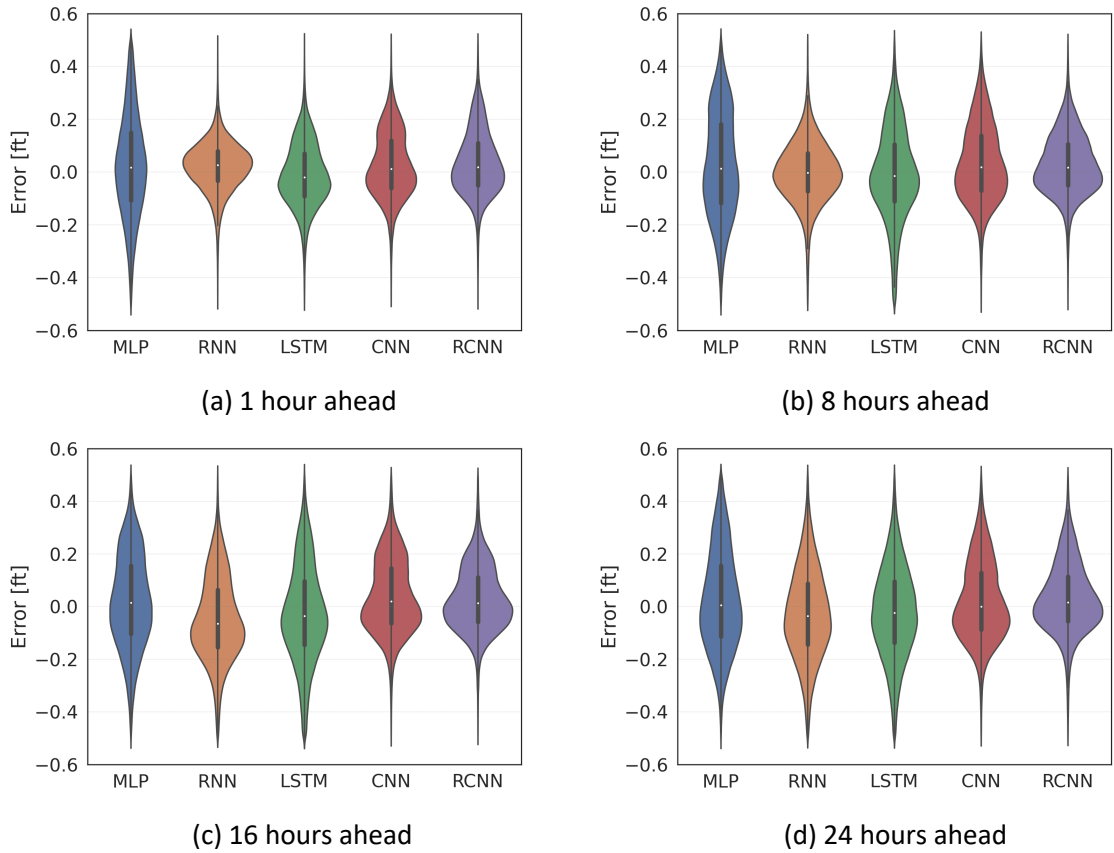

*Figure 15. Violin plots showing density distributions of actual errors for the five DL methods at all locations with forecast lead time: (a) 1 hour; (b) 12 hours; (c) 18 hours; and (d) 24 hours. Statistically significant difference among these violin plots can be observed by shrinking the range of prediction errors.*

**C2.2 Violin plots expanded over location**

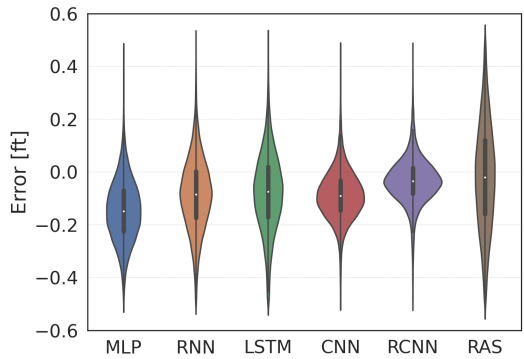
(a) S1

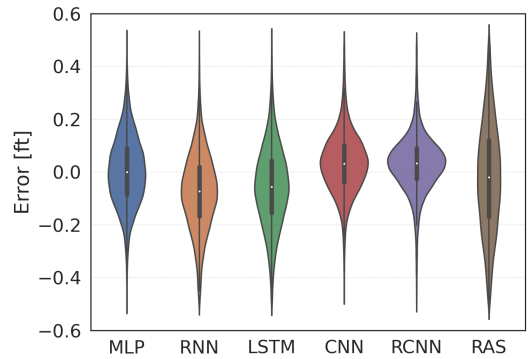
(b) S25A

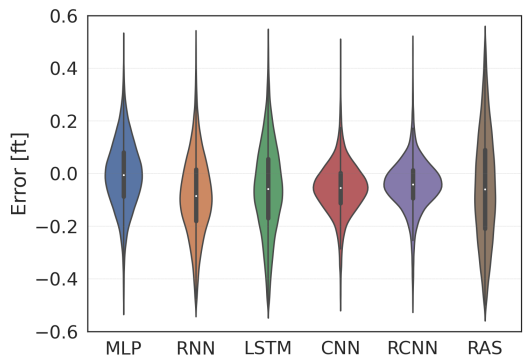
(c) S25B

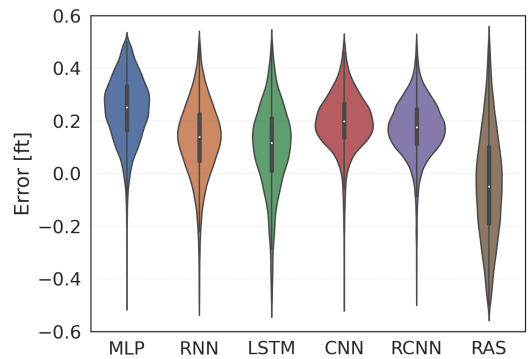
(d) S26

*Figure 16. Violin plots of error distributions for 24-hours-ahead forecasts at each of four locations, S1, S25A, S25B, and S26. Statistically significant difference among these violin plots can be observed by shrinking the range of prediction errors.*

**C3. Violin plots with 20% noise added**

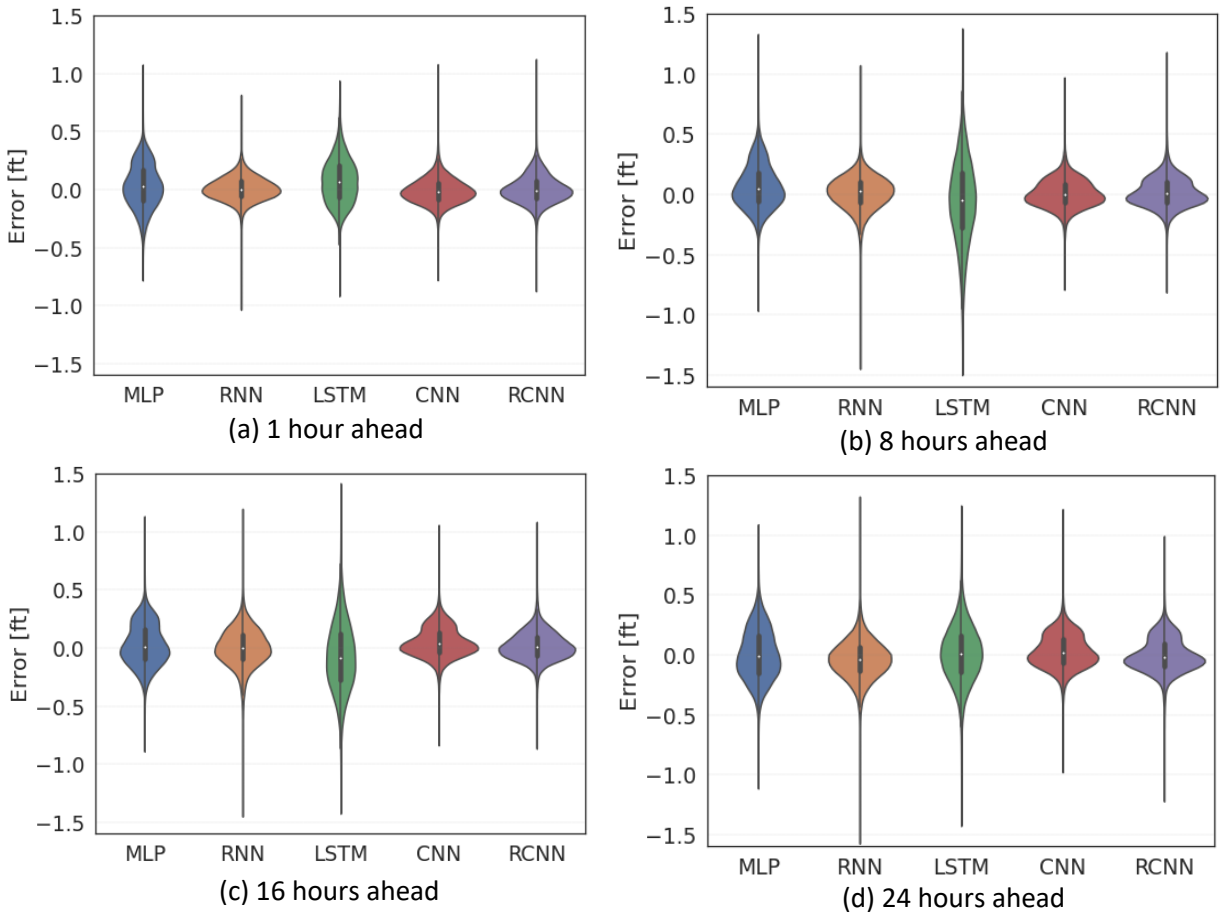

Figure 17. Violin plots with 20% noise added for the five DL methods at all locations with forecast lead time: (a) 1 hour; (b) 12 hours; (c) 18 hours; and (d) 24 hours. We zoomed in by magnifying the y-axis of the violin plots for clear comparison.

## C4. Observed and predicted water stages at other locations

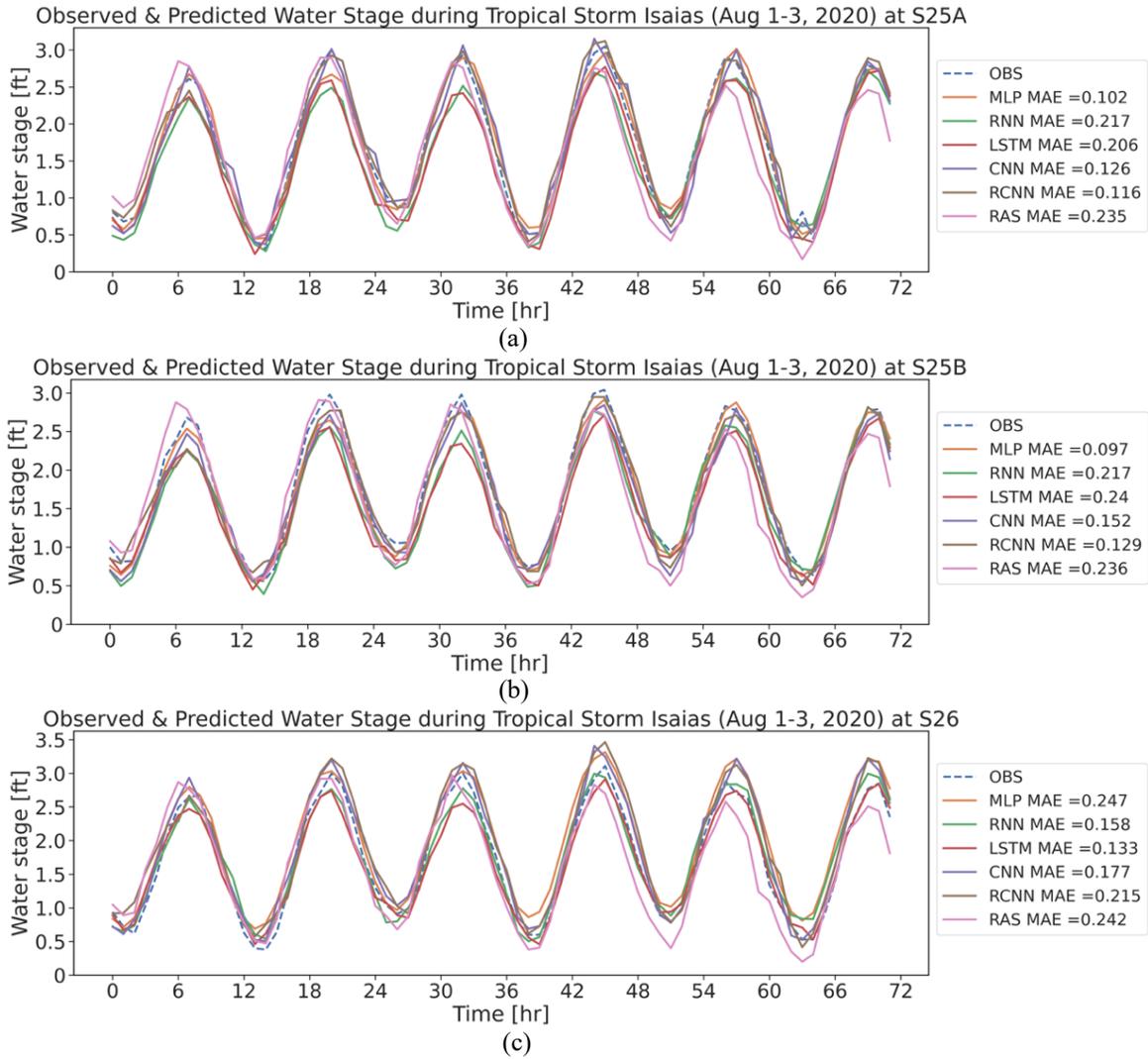

*Figure 18. Observed and Predicted Water Stages during Tropical Storm (2020): (a) S25A; (b) S25B; and (c) S26.*

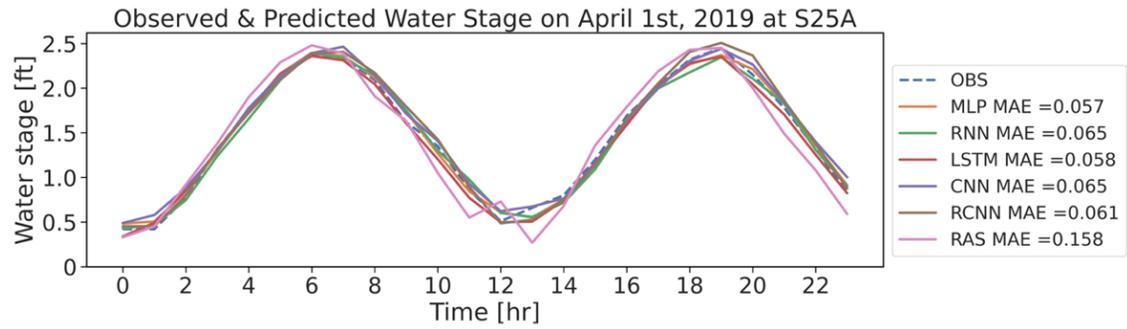

(a)

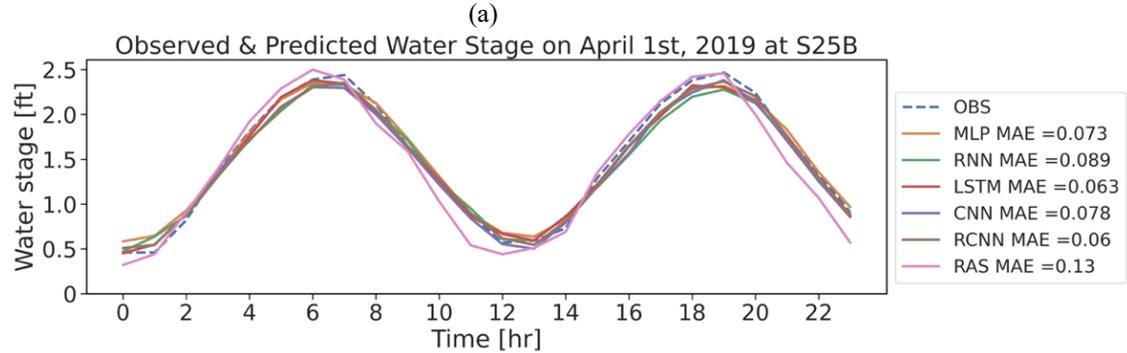

(b)

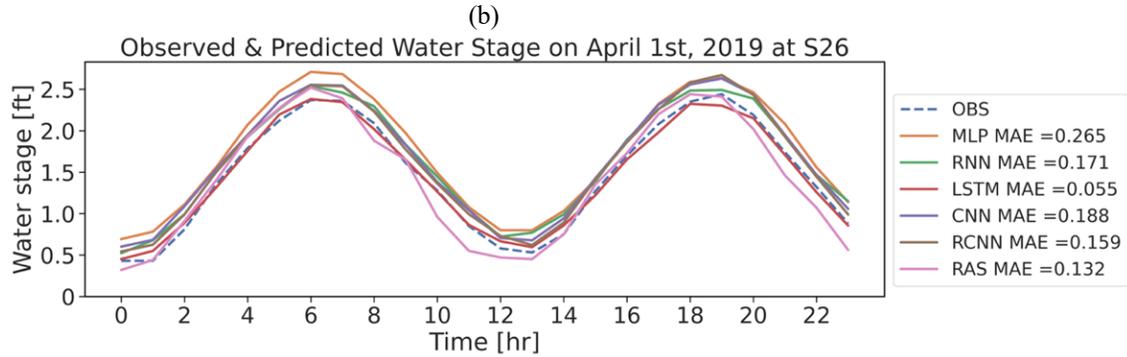

(c)

*Figure 19. Observed & Predicted Water Stage on April 1, 2019: Water Stage at (a) S25A; (b) S25B; and (c) S26.*

## D. Statistical Significance of model comparisons

### D1. P-values for Wilcoxon tests between RCNN and other DL models

*Table 8. P values between RCNN and other models (see Figure 4)*

|      | RCNN_MLP | RCNN_RNN | RCNN_LSTM | RCNN_CNN | RCNN_RAS |
|------|----------|----------|-----------|----------|----------|
| T+1  | 1e-21    | 4e-109   | 0         | 4e-73    | 0        |
| T+8  | 2e-18    | 0        | 0         | 1e-57    | 0        |
| T+16 | 2e-8     | 0        | 0         | 0        | 0        |
| T+24 | 1e-27    | 0        | 0         | 0        | 0        |

### D2. P-values for Wilcoxon tests between DL models (24-hour-ahead prediction) and HEC-RAS at different locations

*Table 9. P values between DL models and HEC-RAS at different locations (see Figures 5 and 14)*

|      | MLP_RAS | RNN_RAS | LSTM_RAS | CNN_RAS | RCNN_RAS |
|------|---------|---------|----------|---------|----------|
| S1   | 0       | 2e-180  | 4e-165   | 9e-252  | 5e-10    |
| S25A | 1e-39   | 4e-102  | 4e-38    | 2e-182  | 4e-181   |
| S25B | 2e-165  | 7e-21   | 2e-10    | 3e-9    | 2e-35    |
| S26  | 0       | 0       | 0        | 0       | 0        |